\documentclass[conference]{IEEEtran}
\IEEEoverridecommandlockouts
\usepackage{cite}
\usepackage{amsmath,amssymb,amsfonts}
\usepackage{algorithm,algorithmic,amsmath}
\usepackage{graphicx}
\usepackage{textcomp}
\usepackage{xcolor}

\usepackage{lettrine}
\usepackage{wrapfig}
\usepackage{lipsum}
\usepackage{subfig}
\usepackage{booktabs}
\usepackage{multirow}
\usepackage{hhline}

\usepackage{hyperref}
\hypersetup{draft}
\usepackage{varioref}
\usepackage{comment}
\usepackage{array}
\usepackage{float}
\usepackage{placeins}

\def\BibTeX{{\rm B\kern-.05em{\sc i\kern-.025em b}\kern-.08em
    T\kern-.1667em\lower.7ex\hbox{E}\kern-.125emX}}

\def\BibTeX{{\rm B\kern-.05em{\sc i\kern-.025em b}\kern-.08em
    T\kern-.1667em\lower.7ex\hbox{E}\kern-.125emX}}
\providecommand{\keywords}[1]
{
  \small	
  \textbf{\textit{Keywords---}}\textbf{\textit{#1}}
}

\makeatletter

\def\ps@IEEEtitlepagestyle{%
    \def\@oddfoot{\mycopyrightnotice}%
    \def\@evenfoot{}%
}
\def\mycopyrightnotice{%
    {\footnotesize  {\fontsize{8}{12}\fontfamily{ptm}\selectfont 978-1-5386-7097-2/18/\$31.00 \copyright 2018 IEEE\hfill }}
    \gdef\mycopyrightnotice{}
}



\makeatletter
\newcommand*\titleheader[1]{\gdef\@titleheader{#1}}
\AtBeginDocument{%
  \let\st@red@title\@title%
  \def\@title{%
    \bgroup\normalfont\large\centering\@titleheader\par\egroup
    \vskip1.5em\st@red@title}
}
\makeatother

\title{Towards Human Pulse Rate Estimation from Face Video: Automatic Component Selection and Comparison of Blind Source Separation Methods}
\titleheader{ \raggedright {\fontsize{8}{12}\fontfamily{ptm}\selectfont 2018 International Conference on Intelligent Systems (IS) }}

\begin{document}

\author{
\IEEEauthorblockN{Vladislav Ostankovich\textsuperscript{*} \thanks{ * These two authors contributed equally}}
\IEEEauthorblockA{\textit{Institute of Robotics}
 \\ \textit{Innopolis University} \\ Innopolis, Russia \\
v.ostankovich@innopolis.ru}
\and
\IEEEauthorblockN{Geesara Prathap\textsuperscript{*} }
\IEEEauthorblockA{\textit{Institute of Robotics}
 \\ \textit{Innopolis University} \\ Innopolis, Russia \\
g.mudiyanselage@innopolis.ru}
\and
\IEEEauthorblockN{Ilya Afanasyev}
\IEEEauthorblockA{\textit{Institute of Robotics}
 \\ \textit{Innopolis University} \\ Innopolis, Russia \\
i.afanasyev@innopolis.ru}}
\maketitle

\begin{abstract}
Human heartbeat can be measured using several different ways appropriately based on the patient condition which includes contact base such as measured by using instruments and non-contact base such as computer vision assisted techniques. Non-contact based approached are getting popular due to those techniques are capable of mitigating some of the limitations of contact-based techniques especially in clinical section. However, existing vision guided approaches are not able to prove high accurate result due to various reason such as the property of camera, illumination changes, skin tones in face image, etc. We propose a technique that uses video as an input and returns pulse rate in output. Initially, key point detection is carried out on two facial subregions: forehead	 and nose-mouth. After removing unstable features, the temporal filtering is applied to isolate frequencies of interest. Then four component analysis methods are employed in order to distinguish the cardiovascular pulse signal from extraneous noise caused by respiration, vestibular activity and other changes in facial expression. Afterwards, proposed peak detection technique is applied for each component which extracted from one of the four different component selection algorithms. This will enable to locate the positions of peaks in each component. Proposed automatic components selection technique is employed in order to select an optimal component which will be used to calculate the heartbeat. Finally, we conclude with comparison of four component analysis methods (PCA, FastICA, JADE, SHIBBS), processing face video datasets of fifteen volunteers with verification by an ECG/EKG Workstation as a ground truth. \\
\end{abstract}


\section{Introduction}
The human pulse is a rhythmic oscillation of the vessels that correspond to the contractions of the heart and it is one of the most important indicators that helps to track whether everything is good with the heart. One can understand what it is worth to be afraid of by observation of pulse accelerations and slowdowns. That is why, according to their frequency, they judge the state of the heart muscle. So, the pulse can be characterized by the strength and rhythm of the heartbeat and even the state of the vessels through which blood flows.
To simply measure the pulse rate one can use special devices like a wrist pulsometer. You can even measure the pulse yourself without any instruments, putting your fingers in the place where the arteries go close to the skin and counting the number of strokes over time. To discover heart diseases the cardiologists use electrocardiographs (ECG). All these methods imply contact based measurements, however some people may have easy damageable skin or impossibility of going to the doctor to record the ECG. In these cases people need to have possibility of being monitored remotely and without contact with any devices \cite{kranjec2014}.

The computer vision systems help to overcome this problem on a completely new level - now we can monitor the human pulse rate without contact equipment that be very useful for self-diagnosis, remote medical consultations and patient monitoring in hospitals \cite{kranjec2014,sikdar2016}.
In the last decade, some studies \cite{verkruysse2008,lewandowska2011,Poh:10,wu,balakrishnan,tarassenko2014,tulyakov2016} have shown that human pulse rate signals can be acquired remotely from a standard RGB camera by processing a video dataset of a patient face under conventional illumination conditions.
In 2010-2013, in some MIT papers \cite{Poh:10,wu,balakrishnan} methods that allow to measure the human pulse from a video, enhancing the smallest facial color changes associated with blood inflow and outflow during heartbeat, were proposed. These investigations inspired the authors of this paper to contribute and update this methodology by automation of component analysis and comparison of various blind source separation methods for estimating human pulse from head motions in video.
This methodology allows to measure the pulse, even if the video is very noisy, a person stands with his back to the camera or wears a mask. The new algorithm doesn't use the color changes of the face, instead it tracks micro motions of head caused by tremors of blood flowing through arteries. With each beat of the pulse, enough blood enters into the brain, so that the head swings slightly - these movements are completely involuntary. The accuracy of such pulse measurement method is comparable to an electrocardiograph.

There are a few other studies on heart beat estimation from video. In \cite{pal2013robust} authors discovered the heart beat using the Iphone camera with turned on flashlight and fingertip covering both. They monitored the variation in mean red channel value and used Mealy Machine to analyze its frequency in consecutive input video frames.
Authors of \cite{irani2014improved} improved the algorithm described in \cite{balakrishnan} using discrete cosine transform (DCT) instead of Fast Fourier Transform (FFT) to analyze the periodicity of signals after principal component analysis (PCA). 
The method described in \cite{Poh:10} is ``based on automatic face tracking
along with blind source separation of the color channels into independent
components''.

In our work, we try out different component analysis techniques to deal with the Blind Source Separation (BSS) of motion corresponding to pulse from extraneous motions. After extracting independent components by using BSS techniques, next challenging problem is to detect pulse rate. In order to do it, the optimal independent component should be properly identified either manually or automatically. When concerning about practical side of this, manual approach is not feasible. Thus, it needs to be selected automatically. In \cite{wedekind2018robust} authors proposed automated component selection routines based on heartbeat detections. Furthermore, their approach is compared with standard concepts such as higher order moments and frequency-domain features for automatic component selection by validating on ECG dataset which consists of healthy subjects performing a motion protocol and the MIT-BIH Arrhythmia Database \cite{moody1992bih}. The paper \cite{iwai2017automatic} proposes an automatic component selection method for noise reduction in magnetocardiograph (MCG) signals which utilizes the autocorrelation function for the extracted components by the use of ICA. In comparison to ECG signals, component selection with MCG signals is challenging due to background magnetic noise because MCG signals are extremely small compared to them.    

Our study is completely different from the studies in \cite{wedekind2018robust} and \cite{iwai2017automatic} because we are trying to detect heartbeat indirectly by observing micro facial expression. Hence, common properties of ECG or MCG signals cannot be incorporated for automatic components selection process. However, there should exist a repetitive pattern on either one of the extracted independent components which can be correlated with heartbeat estimation automatically. Our investigation is to find out how those micro-movements of the face can be used to estimate the human heartbeat.

\textbf{Our contributions}
\begin{enumerate}
\item Performing comparative study for four different Component Analysis methods (PCA, FastICA, JADE and SHIBBS).
\item Developing a peak detection technique which incorporates Moving Dynamic Time Warping (MDTW) which helps to transform into another spatial domain where peak detection is quite simplified.
\item Proposing an Automatic Component Selection (ACS) technique for selecting an optimal component. Here simple statistical property ''skewness'' is employed on differences between adjacent detected peak points after removing bad components which consists of high variance (more than $3\sigma$, where $\sigma$ denotes the stand deviation). 
\item Collected and published dataset with videos and ECG signals \cite{github_dataset}.
\end{enumerate}

\section{Methodology}
Proposed method uses video as an input and returns pulse rate as an output. We start with a detection of person's face on the first frame. Then we detect and track facial feature points throughout the whole video. Only vertical displacement of feature points is considered. Next we interpolate the data vector to match sampling frequency of ECG device used in the experiment and remove unstable features. Afterwards, the temporal filtering is applied to isolate frequencies of interest. Then we use component analysis methods to separate cardiovascular pulse signal from other causes of motion, that could be caused by external factors. We apply four component analysis algorithms for comparative study. Then after applying proposed Peak Detection (PD) technique, it will enable us to locate the positions of peaks in each component. Proposed Automatic Components Selection (ACS) technique is employed in order to select an optimal component which is used to calculate the heart rate by incorporating the PD results. Following subsections will explain each part of the proposed methodology in detail.

\begin{figure*}[!ht]
\begin{center}
\includegraphics[width=\textwidth]{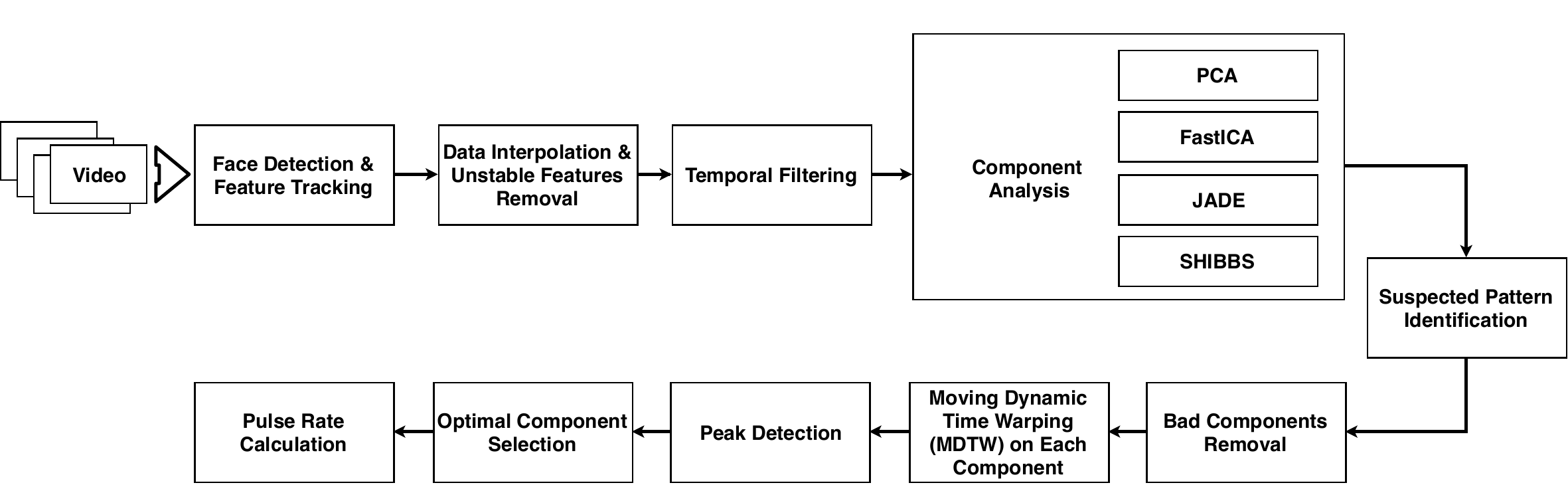}
\caption{\label{f:project_schema} {\fontsize{8}{12}\fontfamily{ptm}\selectfont Sequence of steps of proposed methodology how pulse rate is calculated from a facial video}}
\end{center}
\end{figure*}

\subsection{Preprocessing}
The preprocessing is the initial stage where it is needed to make some observations from video before using them in component analysis. This includes several steps such as person's face detection and feature points tracking, data interpolation, unstable features removal and temporal filtering. The approach is described in \cite{balakrishnan}, but in our work a few minor changes are made.

\subsubsection{Feature Tracking}
\begin{figure}[!ht]
\begin{center}
\includegraphics[scale=0.35]{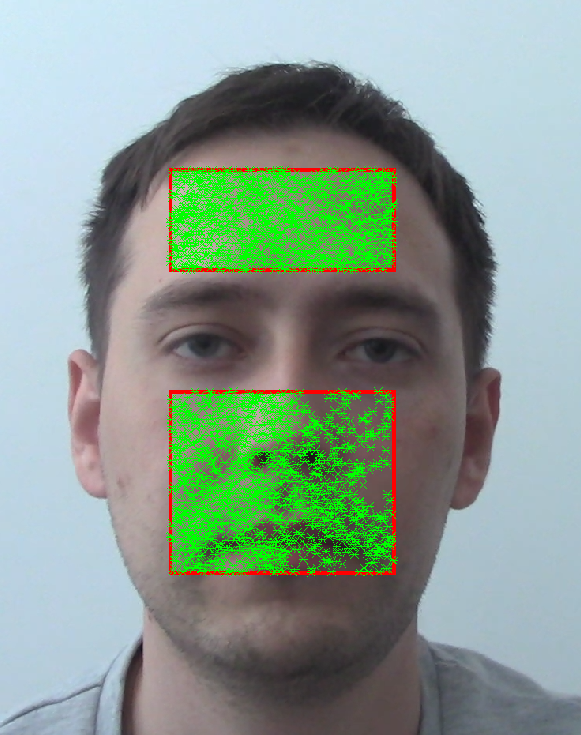}
\caption{\label{f:facenfeatures} {\fontsize{8}{12}\fontfamily{ptm}\selectfont Forehead and nose-mouth facial subregions with corresponding feature points.}}
\end{center}
\end{figure}

First of all, there is a need to detect the face on a video. For this purpose, the Viola-Jones algorithm \cite{Viola2001} was applied. As indicated in \cite{balakrishnan} the face bounding box should be adjusted to be sure that all features will be located inside the face region. Therefore we used middle 50\% width and 90\% height from top of initial bounding box. Then we need to get rid of eye regions due to the reason that the blinking will negatively affect future results. Consequently, the face region was divided into two subregions, namely forehead and nose-mouth regions. The forehead region is spanning 25\% height from top and the nose-mouth region is spanning 55\% from bottom, therefore the eye region spanning from 25\% to 55\% height from top was removed. Two subregions are shown in Fig. \ref{f:facenfeatures}.
To track feature points of these regions we used Kanade-Lucas-Tomasi algorithm \cite{tomasi1991detection}. Each feature point was tracked for the whole duration of video thus obtaining its location $x$ and $y$ at each frame. As proposed by \cite{balakrishnan} only the vertical components were used in our analysis as well. The feature points for two subregions are also shown in Fig. \ref{f:facenfeatures}
~

After all we get the matrix $Y$ containing the vertical tracking position for all features during the whole video. The size of matrix $Y$ is $F \times T$, where $F$ is a number of all features and $T$ depicts the number of frames. 

\subsubsection{Data Interpolation \& Unstable Features Removal}
Next we need to interpolate our data matrix to fit the ECG model  used as the ground truth. Since the ECG device's sampling rate was set to 250 Hz and the video camera's frame rate was 25 fps, we needed to interpolate the data in 10 times. The cubic spline interpolation algorithm was applied \cite{schoenberg1988contributions}. Thus, the new matrix $Y_{interp}$ has the size $F \times 10T$.

After the interpolation we need to get rid of unstable features. As described by \cite{balakrishnan} ''To retain the most stable features we find the maximum distance (rounded to the nearest pixel) traveled by each point between consecutive frames and discard points with a distance exceeding the mode of the distribution''. The resultant matrix $Y_{stable}$ is a matrix containing only stable features from $Y_{interp}$. 


\subsubsection{Temporal Filtering}
The temporal filtering is applied further since we don't need all frequencies to be present. Since we are interested in cardiovascular pulse, we need to consider the frequencies of 1-2 Hz (that corresponds to 60-120 beats per minute). The authors in \cite{balakrishnan} chose the frequencies from 0.75 to 5 Hz for their experiment. We decided to follow their proposal. Moreover we used the same temporal filter, that is $5^{th}$ order Butterworth one. After temporal filtering of $Y_{stable}$ matrix we obtained new matrix $Y_{filtered}$ of same size.

\subsection{Component Analysis}

Having hundreds of feature points extracted from face and tracked during the whole video we need to observe what type of signal(s) they could describe together. Since we are interested in obtaining the cardiovascular pulse, we need to separate (isolate) it from other causes of feature point motions that could intervene. Such extraneous motions could be caused by respiration, facial expression changes or vestibular activity \cite{balakrishnan}. The decomposition into subsignals can be achieved by different component analysis (CA) techniques. Principal Component Analysis (PCA) was applied in \cite{balakrishnan}, where the authors found that there is no reason to consider eigenvectors beyond the fifth one. Thus, the number of subsignals found is five as well. We decided to stick to this number, and only first five signals were considered in PCA and in all of the following CA methods. 

Considering the input data matrix $Y_{filtered}$ with its number of features being $N$ and time vector described by $T$, any of proposed CA methods would decompose a matrix of size $N \times T$ into $5 \times T$, where the five represents the number of extracted subsignals. The following CA methods were used in the experiment:

\subsubsection{PCA}

Principal Component Analysis is one of the main ways to reduce the data dimensionality by losing the least amount of information \cite{abdi2010}. Strictly speaking, this method approximates the n-dimensional cloud of observations to an ellipsoid (also n-dimensional), whose semiaxes will be future main components. And when we project to such axes (decrease in dimension), the greatest amount of information is saved. This method allows you to reduce the number of variables by selecting the most volatile ones. Namely, in its mathematical essence - it's just a change of variables, a transition to new ones. The method is applied in many areas, for example in pattern recognition, computer vision or data compression. The calculation of the main components is reduced to the calculation of eigenvectors and eigenvalues of the covariance matrix of the original data or to the singular decomposition of the data matrix.

\subsubsection{FastICA}

This ICA method was proposed as a method for solving the problem of Blind Separation of Signals or Blind Source Separation, that is, the separation of independent signals from pre-mixed data \cite{comon, hyvarinen99b,hyvarinen00,hyvarinen01}. FastICA is an effective and popular algorithm for independent component analysis, invented by Aapo Hyvärinen at the Helsinki Polytechnic University \cite{bingham,hyvarinen99}. ``The algorithm is based on the iterative scheme of a fixed point, maximizing non-Gaussianity as a measure of statistical independence'' \cite{apolloni2011neural}. Before the FastICA algorithm can be applied, input vector data must be centered and whitened.

\subsubsection{JADE}

The JADE (Joint Approximate Diagonalization of Eigenmatrices) is an independent component analysis algorithm that obtains source signals after separation of mixture signal. It searches for a rotation matrix that collectively diagonalizes its own matrices derived from fourth-order moments after data whitening. The fourth order moments are non-Gaussianity measures, that are utilized as an intermediary for characterizing independence between the source signals. Good statistical performance is achieved by involving all cumulants of the 2nd and 4th order and rapid optimization is achieved through the use of the method of joint diagonalization \cite{cardosoJade}.

\subsubsection{SHIBBS}

One more signal separation technique is SHIBBS (Shifted Block Blind Separation). The author of \cite{cardosoShibbs} stated that both JADE and SHIBBS separate the source signals similarly. Additionally the authors of \cite{miettinen13} showed that for high-dimension data with large number of features the JADE was not very practical. Moreover, in the studies of the author JADE took much more computational time than SHIBBS.

\section{Peak Detection (PD) \& Automatic Component Selection (ACS)}

\begin{figure}[!ht]
\begin{center}
\includegraphics[scale=0.15]{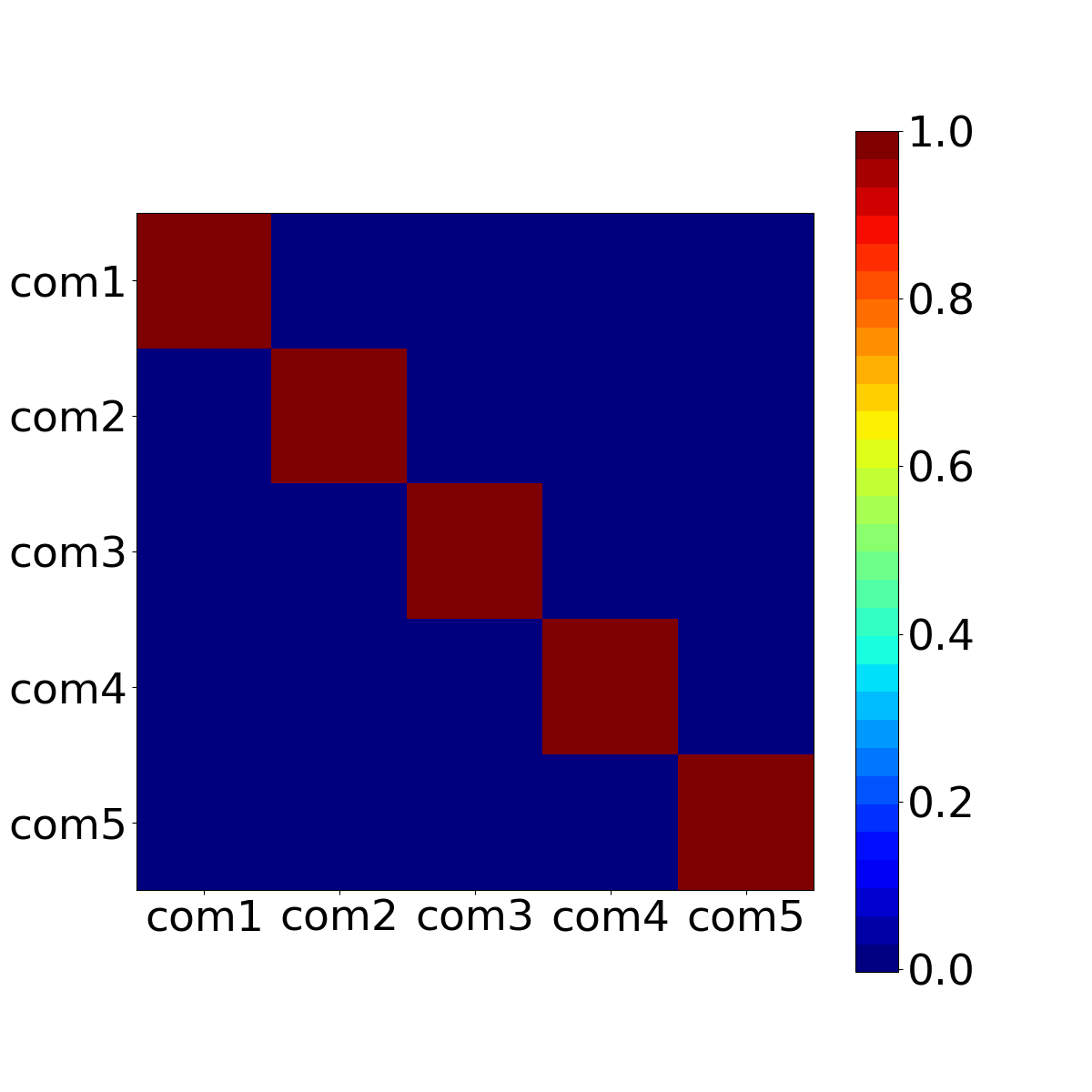}
\caption{\label{f:jade_correlation} {\fontsize{8}{12}\fontfamily{ptm}\selectfont Correlation among the five extracted components after performing JADE algorithm.}}
\end{center}
\end{figure}

After performing either PCA, FastICA, JADE or SHIBBS method we have five extracted independent components. These methods can be considered as blind source separation as well as dimensionality reduction techniques. Thus, the final output will be a set of independent components which can be verified by looking at Fig.~\ref{f:jade_correlation} where there is no correlation between the components. All future steps in this section of the paper will be explained based on the results extracted after applying the JADE algorithm. However the same concept can be applied to any other component extraction method as well.

\begin{figure}[!ht]
\begin{center}
\includegraphics[width=\linewidth]{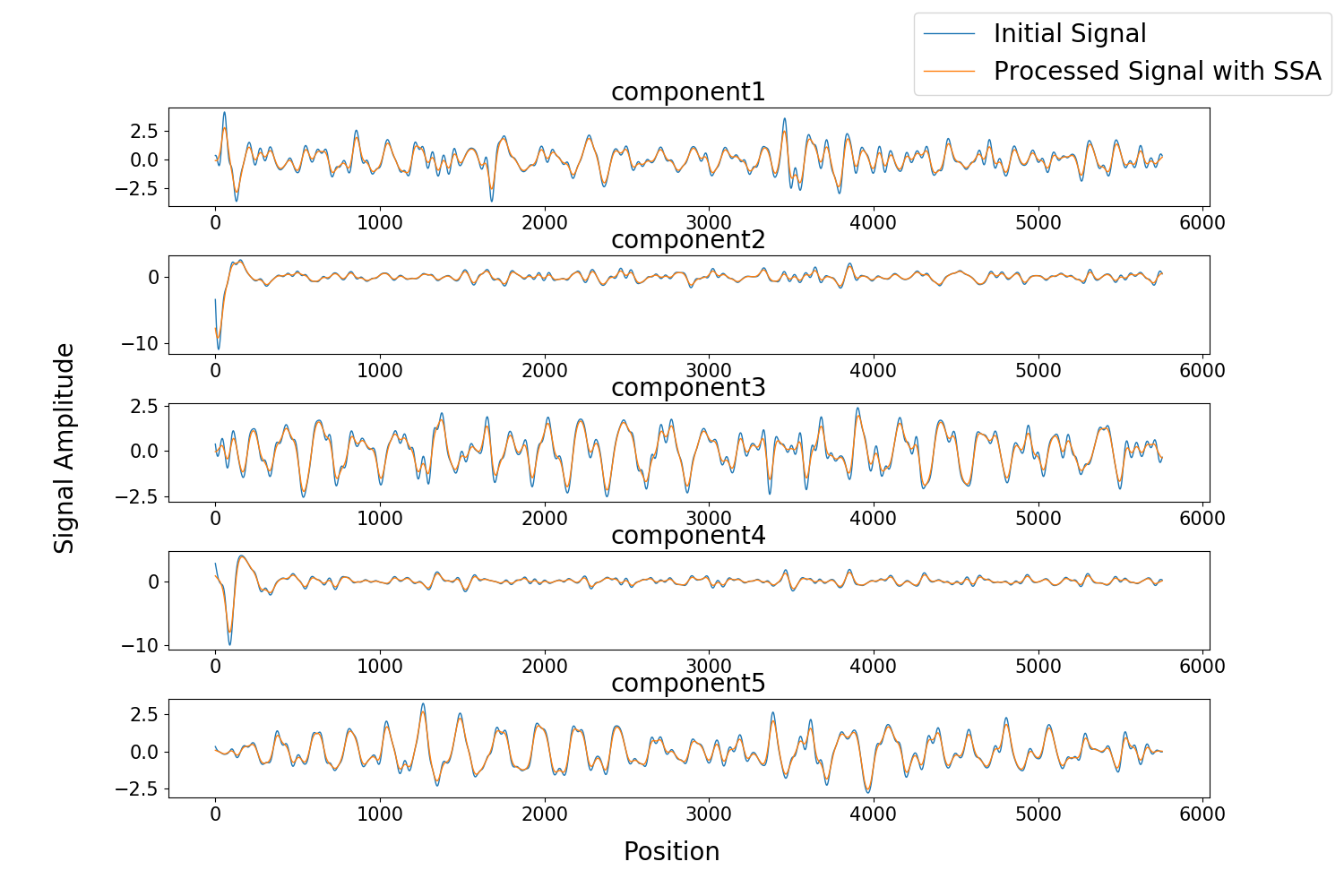}
\caption{\label{f:withandwithoutssa} {\fontsize{8}{12}\fontfamily{ptm}\selectfont Normalized extracted components from JADE algorithm and the result of applying the SSA on each component.}}
\end{center}
\end{figure}

The accuracy of the ACS highly depends on the smoothness of signals. In other words, the main direction or prominent principle components should be extracted for a given component which can be used to represent the selected signal without loosing its information. This process will reduce the noise if there are any. However, this process is not needed to be executed because the input to the ACS is the output from component analysis algorithm where it gives independent components. But this can be verified by applying Singular Spectrum Analysis (SSA)~\cite{golyandina2013singular} on each component. SSA is a nonparametric spectral estimation technique which can be applied for a given time series. As an example, a time series can be defined as $X=(x_1,...,x_N)$ of length N where N should be greater than 2 and X should be a non zero real value. SSA consists of two interdependent stages: decomposition and reconstruction where in the decomposition stage a provided time series is decomposed by using Singular Value Decomposition (SVD)~\cite{de1994singular} and in the reconstruction stage it is disjointed into subsets (principal components) by using eigentriple grouping~\cite{golyandina2013singular}. The basic idea of the SSA is to find out the main direction of a given time series. If the 1st principal component is selected it will give the main direction. When cascading all the principal components it will give the original time series. Since the main idea of applying SSA to extracted signal components is to verify the hypothesis which was made at the beginning of this paragraph. So it can be verified by looking at Fig.~\ref{f:withandwithoutssa} which shows the result after applying the SSA by cascading first 3 principal components which SSA provides. 
%
%
%
%
%
%
%
%
%
%
%
%

\subsection{Suspected Pattern Identification}

\begin{figure}[!ht]
\begin{center}
\includegraphics[width=\linewidth]{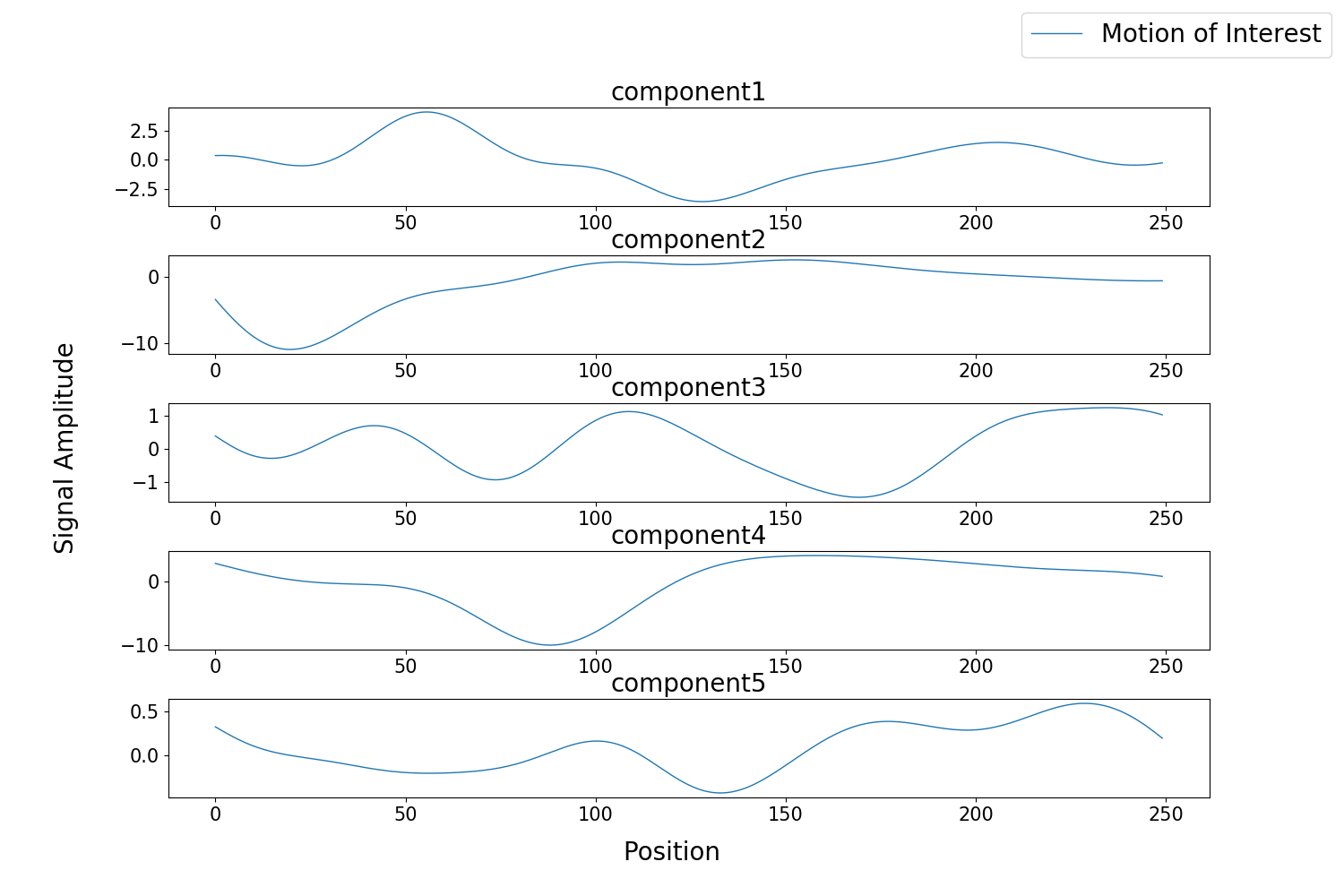}
\caption{\label{f:motion_of_interest}{\fontsize{8}{12}\fontfamily{ptm}\selectfont Motions of interest corresponding to each component.}}
\end{center}
\end{figure}

In the component extraction stage, five suspected components are decomposed. To select an optimal component among them, it is needed to identify the motion of interest which should be unique to each extracted component. The motion of interest also should be repeatable because heartbeat itself contains a repetitive pattern. However, the pattern which belongs to each component cannot be defined beforehand because it varies from person to person as well as what kind of actives he or she is undergoing at that moment. Thus, a pattern corresponding to each component is chosen from the respective extracted component itself by element-wise aggregation(average) of values of three samples corresponding to 2nd, 8th and 16th seconds. In Fig.~\ref{f:motion_of_interest}, the extracted motions of interest corresponding to each component are shown.

\subsection{Bad Components Removal}

As shown in the Fig.~\ref{f:withandwithoutssa}, it can be clearly seen that the beginning of the 2nd and 4th components consists of high variance compared to the latter part of the same signal. This behavior is due the affect of extracting independent components which can not be eliminated beforehand. Subsequently, this will badly affect the ACS and PD as well. Thus, to remove the high variance components, following condition (see Algorithm \ref{a:detect_bad_component}) is used to identify whether a considered component is bad or good.
%
%
%
%

\begin{algorithm} 
  \caption{{\fontsize{8}{12}\fontfamily{ptm}\selectfont Removing bad components if its max or min is grater or less than 3$\sigma$ and -3$\sigma$ respectively}}
  \label{a:detect_bad_component}
  \begin{algorithmic}[1]
      \FOR{$component_i \in \{1,\dots,5\}$}
        \STATE $min\_value = min(component_i)$
        \STATE $max\_value = max(component_i)$
        \STATE $overfitting\_value = 3*\sigma(component_i)$
        \IF{$|int(min\_value + overfitting\_value)|> 0 \quad OR \quad |int(max\_value - overfitting\_value)|> 0$}
          \STATE Consider as a bad component
         \ENDIF 
      \ENDFOR
  \end{algorithmic}
\end{algorithm}
	
\subsection{Applying proposed Moving Dynamic Time Warping (MDTW) approach on each component}

Dynamic Time Warping (DTW)~\cite{keogh2005exact} is a well-known technique for an optimal way of aligning two temporal sequences. For instance, in the article \cite{prathap2018} MDTW was applied for retrieving a suspected motion of interest from raw signals.
The main objective for MDTW is to wrap two sequences in a nonlinear passion where they may vary in speed for measuring the similarity between each other. Lets define two (time-dependent) temporal sequences: $X=(x_1,...,x_N)$ and $Y=(y_1,...,y_M)$ where $N \in N$ and $M \in N$ are the length of each signal respectively. In the DTW, two sequences are considered as two different feature sets where it tries to find a local cost measurement $c(X, Y)$ by using one of the distance measurement methods such as the Manhattan distance~\cite{black2006manhattan}  which is used in our proposed solution. If $c(X, Y)$ is very small or low cost it indicates that there is a high similarity between $X$ and $Y$. On the other hand, if the cost is very large then there will be a high variation between $X$ and $Y$.
%
%
%
%

In the proposed MDTW, the following approach is employed. Window size ($W$) is equal to the size of motion of interest ($X$). Lets denote $Y$ as the extracted signal corresponding to a selected motion of interest. Initially, DTW is applied between $X$ and the first portion of $Y$ which should be equal to the size of $X$. The window is moved by a fixed number of points (e.g. 5), then apply the DTW between a new portion of $Y$ and $X$. This process is carried out until there is no portion that can be separated out from $Y$. Each time when DTW is applied, the corresponding position in $Y$ and the distance between two components are stored in two vectors, and lets depict them as the position vector ($Pos_v$) and distance vector ($Dis_v$) respectively.

\begin{table*}[t!]
 \captionsetup{font=scriptsize}
  \centering
  
  \caption{ {\fontsize{8}{12}\fontfamily{ptm}\selectfont \textsc{ COMPARISON BETWEEN PULSE RATE CALCULATION BY PROPOSED METHOD AND GROUND THROUGH VALUE IN BPM (BEATS PER MINUTE). }}}
   \begin{tabular}{|c|c|c|>{\bfseries}c|c|>{\bfseries}c|c|c|>{\bfseries}c|c|>{\bfseries}c|c|>{\bfseries}c|}
    \hline
    \multirow{2}{*}{PersonId} & \multicolumn{5}{c|}{Normal Condition (bpm)} & \multicolumn{5}{c|}{Undergoing Some Physical Activity (bpm)} \\
    \hhline{~----------}
    & FastICA & PCA & JADE & SHIBBS & GT & FastICA & PCA & JADE & SHIBBS & GT \\
    \hline
    \multirow{1}{*}{P1} & 76.42&65.54&70.16&68.87&64.71&42.86&76.49&81.92&79.25&72.97\\
    \hhline{-----------} 
    \hline
    \multirow{1}{*}{P2} &39.39&72.96&79.43&61.48&78.0&59.89&74.93&72.18&68.83&76.62\\
     \hhline{-----------} 
    \hline
    \multirow{1}{*}{P3} & 86.57&64.55&67.65&62.01&58.61&20.23&62.87&62.69&59.17&60.61\\
     \hhline{-----------} 
    \hline
    \multirow{1}{*}{P4} & 65.28&68.12&79.33&72.32&87.09&24.24&60.91&65.54&24.29&105.37
\\
     \hhline{-----------} 
    \hline
    \multirow{1}{*}{P5} & 50.45&75.45&72.05&67.19&78.12&30.47&67.69&76.84&61.48&90.72
\\
     \hhline{-------------} 
    \hline
    \multirow{1}{*}{P6} & 39.26&70.19&68.61&64.42&62.62&64.21&69.98&76.61&64.22&92.36
\\
    \hhline{-------------} 
    \hline
    \multirow{1}{*}{P7} &60.34&67.06&70.42&66.99&73.14&42.81&83.42&96.1&59.7&87.26
\\
     \hhline{-----------} 
    \hline
    \multirow{1}{*}{P8} &59.01&61.92&74.23&63.38&98.06&86.03&91.28&92.16&58.58&98.1
\\
     \hhline{-----------} 
    \hline
    \multirow{1}{*}{P9} &65.93&58.14&60.06&60.69&53.18&62.19&61.4&65.28&74.78&57.2
\\
     \hhline{-----------} 
    \hline
    \multirow{1}{*}{P10} &51.52&39.59&76.55&53.57&82.67&37.38&70.31&65.28&69.84&90.05
\\
     \hhline{-----------} 
    \hline
    \multirow{1}{*}{P11} &71.22&63.77&65.56&70.19&64.17&66.4&79.49&73.53&66.86&70.44
 \\
    \hhline{-----------} 
    \hline
    \multirow{1}{*}{P12} & 53.46&57.99&58.88&61.86&54.49&52.94&61.29&66.8&71.06&58.82
 \\
    \hline
    \multirow{1}{*}{P13} &77.88&73.62&69.62&67.83&65.97&74.61&70.04&82.4&63.25&65.97
\\
     \hhline{-----------} 
    \hline
    \multirow{1}{*}{P14} &47.11&66.2&67.19&69.84&64.52&60.25&53.93&53.46&38.3&95.88 \\
     \hhline{-----------} 
    \hline
    \multirow{1}{*}{P15} & 61.29&46.78&82.9&48.72&78.27&80.83&72.95&63.06&74.46&71.91
 \\
     \hhline{-----------} 
    \hline
  \end{tabular}
  
  \label{t:pulserate}
\end{table*}

\subsection{Peak Detection}

\begin{figure}[!ht]
\begin{center}
\includegraphics[width=\linewidth]{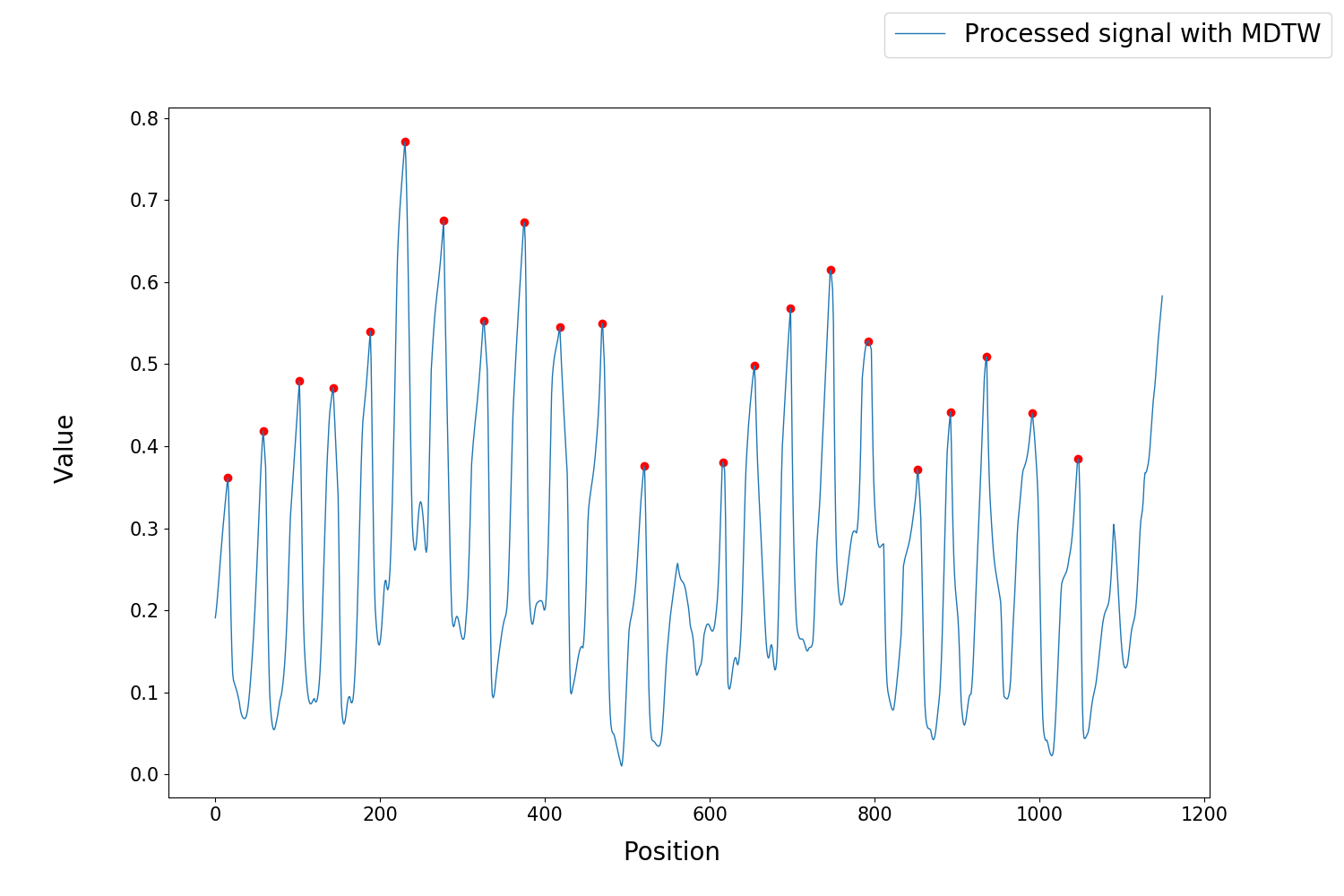}
\caption{\label{f:peakdetection} {\fontsize{8}{12}\fontfamily{ptm}\selectfont  This figure shows the peak detection of the result of the MDTW. The horizontal axis represents $Pos_v$ and the vertical axis represents values of $Dis_v$. A predefined threshold value is used to control the number of detection points. }}
\end{center}
\end{figure}

If peak detection is performed on the extracted signal itself, there will be a quite few numbers of detected peaks that cannot make any useful insight. That is why MDTW is proposed where peak detection is performed in a different space where the result has a proper meaning. Peak detection in this new space is shown in Fig.~\ref{f:peakdetection}. Once peak positions are identified with the help of the $Pos_v$ and $Dis_v$, corresponding positions of an extracted signal component can be identified.
%
%
%
%

\subsection{Optimal Component Selection}

Since there are five components, there should be a way of automatically selecting one component among others which can be used for calculating the pulse rate because there may be some incidents where more than one component will give the same amount of peaks or some of them give unexpected results such as quite big or small number of peaks which cannot be considered. To overcome those issues and select an optimal component the following technique is proposed. 

\begin{enumerate}  
\item Calculate difference between adjacent values of the detected peak points, lets define the output vector of this operation as the $diff\_vector$
\item Calculate the 3rd central moment or skewness of $diff\_vector$ which can be defined as: 
\begin{gather}
 		\frac{1}{(n-1)}\sum_{i=1}^{n-1}(diff\_vector_i-\overline{diff\_vector})^3
 \end{gather} where n is the number of peaks that were detected, $diff\_vector_i$ is each element of $diff\_vector$ and its mean is denoted as $\overline{diff\_vector}$. 
\item Select the component which corresponds to the minimum skewness. This will be the optimal component among others. The skewness is chosen because it is a good indication for determining which measurement of central tendency is the best at finding the 'centre' location.
\end{enumerate}

\begin{figure}[!ht]
\begin{center}
\includegraphics[width=\linewidth]{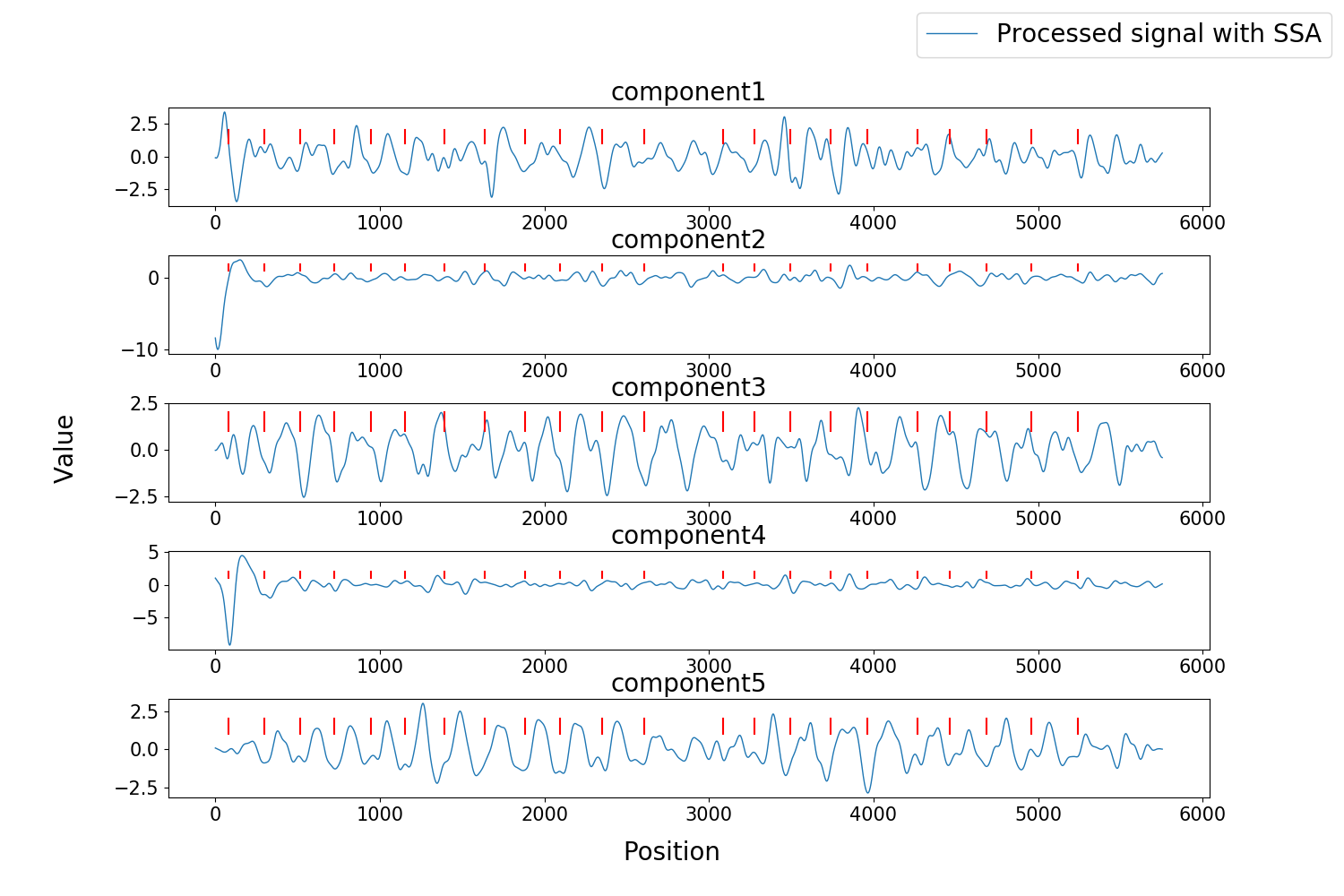}
\caption{\label{f:labeld_component} {\fontsize{8}{12}\fontfamily{ptm}\selectfont The figure shows that locations of peak points which correspond to the component that is chosen form ACS.  In this case, the 5th component was selected and then it is used to locate the peak points of original components which were extracted from the JADE algorithm. }}
\end{center}
\end{figure}

For the selected dataset it gave the fifth component which corresponds with the minimum skewness. Thus, suspected repetitive pattern which corresponds to the fifth component is shown in Fig.~\ref{f:labeld_component}.
%
%
%
%

\subsection{Pulse Rate Calculation}
Pulse rate or beats per minute (bpm) is calculated based on the following formula:
\begin{equation}\label{eq:pulse_detection}
	pulse\_rate = \frac{60}{t_2 - t_1}*N_p 
%
%
\end{equation} where $t_1$ and $t_2$ are the time instances of first and last peak points which were detected by using proposed ACS technique. $N_p$ is the total number of peaks between $t_1$ and $t_2$.

\section{Experimental Procedure}

Testing dataset videos were shot with a Canon Legria HFR 66 camera. 
The ``ECG Lite'' USB cardiograph was used for ground truth comparison \cite{ecglite}. It has 4 ECG clamp electrodes and can be used as ambulatory electrocardiography device for continuously monitoring ECG. The electrodes are placed on wrists and ankles. The device has an adjustable sampling rate, and it was decided to use 250 Hz.

``ECG Control'' software was used to monitor the ECG in real-time and save ECG records to a text file. The proposed method was implemented in MATLAB and Python and available on GitHub. (\footnote{Pulse Detection: \url{https://github.com/vladostan/pulse}\label{pulase_detection}}).

All testing videos have 25 frames per second frame rate. The resolution is Full HD, i.e. 1920x1080 pixels. The average duration of testing video samples is around 20 seconds. Videos were shot indoors. The whole room was evenly illuminated by fluorescent lamps, therefore the persons' faces on videos have no glares or shades. The subjects remained in sitting posture during the footage.

For ground truth data acquisition, subjects were connected to ECG device described in part A. The subjects were examined in two different conditions: in rest and after physical exercises.  This is to detect how the proposed method will cope with the subjects observed at various physical conditions, specifically after the pulse increase.

We decided to publish the dataset on GitHub as well (\footnote{ \url{https://github.com/vladostan/Dataset-for-video-based-pulse-detection}\label{pulase_detection_dataset}}). Collected dataset has 30 videos of 15 persons examined at two different physical conditions (normal and after physical exercises). It also contains ground truth ECG signals that were recorded simultaneously with videos and saved in text files.

\begin{table}[h!]
\captionsetup{font=scriptsize}
\centering
\caption{ {\fontsize{8}{12}\fontfamily{ptm}\selectfont  \textsc{ CALCULATING RMSE (ROOT MEAN SQUARE ERROR) IN BPM FOR TWO PHYSICAL CONDITIONS SEPARATELY WITH RESPECT TO THE COMPONENT ANALYSIS METHOD. }}}
\label{t:comparison}
\begin{tabular}{|c|c|c|}
\hline
Components Extraction Technique & Condition 1 & Condition 2 \\ \hline
FastICA                         &5.88         & 9.47        \\ \hline
PCA                             & 4.60         & 4.88         \\ \hline
\textbf{JADE}  & \textbf{2.07}        & \textbf{4.78}        \\ \hline
SHIBBS                          & 4.09         & 8.12         \\ \hline
\end{tabular}
\end{table}

\section{Experimental Results}

The experiment was conducted with fifteen people to analyze the robustness of the proposed method. Furthermore, to investigate which component extraction algorithm is performed in an optimal way when extracting five independent components from the initial feature matrix that is extracted from the this~\cite{balakrishnan} algorithm. 

The accuracy of the pulse rate estimation depends on the detecting the number of repetitive pattern by using proposed PD technique appropriately. However, PD is depends on feature extraction technique, algorithm which is employed to extracted 5 independent components as well as selecting a proper component by using proposed ACS technique. After peak points are detected, the pulse rate can be calculated using the formula~\ref{eq:pulse_detection}. The results are shown in Table \ref{t:pulserate}.
%
%
%
%

When extracting independent components, JADE is performing well in comparison to PCA, FastICA and SHIBBS which is shown in Table \ref{t:comparison}. Moreover, JADE takes the lowest time to extract independent components from initial feature matrix (see Table \ref{t:ca_time}). This is also another advantage when considering total execution time for the whole process. The execution time also varies from person to person due to different number of tracked facial feature points. When comparing the RMSE in bpm, JADE has lower error rates for the condition 1 compared with condition 2 while scoring the lowest error rate compared to other techniques. The downside of the proposed method is that it is not estimating pulse rate approximately all the time. This can be clearly noticed by looking at the result related to p4 and p8. Since the experiment was undertaken with the same environmental condition, this unexpected behaviour is due to the very low micro-level facial expressions which badly affect the initial feature tracking. Along with that, it will affect on independent component extraction process as well. That is why it has a high error margin between ground truth and estimated values.

\begin{table}[h!]
\captionsetup{font=scriptsize}
\centering
\caption{{\fontsize{8}{8}\fontfamily{ptm}\selectfont  \textsc{ EXECUTION TIME IS TAKEN BY EACH COMPONENT ANALYSIS METHOD TO EXTRACT 5 COMPONENTS FROM FEATURE MATRIX. THE TIME IS SUMMED ALONG TWO PHYSICAL CONDITIONS FOR EACH PERSON. }}}
\begin{tabular}{|c|c|c|>{\bfseries}c|c|>{\bfseries}c|c|c|>{\bfseries}c|c|>{\bfseries}c|c|>{\bfseries}c|}
    \hline
    \multirow{2}{*}{PersonId} & \multicolumn{4}{c|}{Execution time (s)} \\
    \hhline{~----------}
    & FastICA & PCA & JADE & SHIBBS \\
    \hline
    	\multirow{1}{*}{P1} &0.52&0.32&0.08&63.34 \\
    \hline
    	\multirow{1}{*}{P2} &0.37&0.67&0.15&48.49 \\
    \hline
        \multirow{1}{*}{P3} &1.38&1.86&0.41&41.22 \\
    \hline
        \multirow{1}{*}{P4} &0.59&1.63&0.30&45.94 \\
    \hline
        \multirow{1}{*}{P5} &0.63&1.64&0.32&46.94 \\
    \hline
        \multirow{1}{*}{P6} &0.51&1.12&0.25&50.46 \\
    \hline
        \multirow{1}{*}{P7} &2.51&6.61&1.64&47.62 \\
    \hline
        \multirow{1}{*}{P8} &0.47&0.97&0.30&41.34 \\
    \hline
        \multirow{1}{*}{P9} &0.55&1.36&0.27&44.32 \\
    \hline
        \multirow{1}{*}{P10} &0.65&1.51&0.36&36.22 \\
    \hline
        \multirow{1}{*}{P11} &1.05&2.17&0.42&40.75 \\
    \hline
        \multirow{1}{*}{P12} &1.48&4.24&0.96&36.51 \\
    \hline
        \multirow{1}{*}{P13} &0.58&1.43&0.34&34.18 \\
    \hline
        \multirow{1}{*}{P14} &1.19&1.91&0.44&35.27 \\
    \hline
        \multirow{1}{*}{P15} &0.57&1.05&0.26&36.45 \\
    \hline
    	\multirow{1}{*}{Average} &0.87&1.90&0.43&43.27 \\
    \hline
\end{tabular}
\label{t:ca_time}
\end{table}

\section{Conclusion}
Non-contact based heartbeat calculation approaches are getting popular due to the capabilities of mitigating the limitations of contact-based techniques, especially in clinical section. We proposed a method for calculating the pulse rate as a non-contact based approach. The initial stage of the proposed method is to extract five independent components. We have analysed 4 components extraction algorithms and concluded that JADE is the best algorithm among others. Afterwards, positions of peaks of each component were determined by using proposed peak detection technique. Then an optimal component was estimated by employing proposed automatic component selection technique. Finally, the root-mean-square error is calculated to examine how estimated guess varies from the ground truth. Still accuracy of the proposed method is to be improved when the pulse rate is high. We are trying to improve this method which can be run in real time while improving the robustness of the methodology. 

%
%
%
%
\FloatBarrier
\bibliographystyle{ieeetr}
\bibliography{sample}

\end{document}